# A Survey of Current Opportunities for Developing Automated Assessment System for C/C++ Programing Assignments

Muhammad Salman Khan
mskhan@ciitlahore.edu.pk
Department of Computer Science, CIIT, Lahore, Pakistan

Adnan Ahmad
adnanahmad@ciitlahore.edu.pk
Department of Computer Science, CIIT, Lahore, Pakistan

Muhammad Humayoun
dr.humayoun@ucp.edu.pk
Department of Computer Science, UCP, Lahore, Pakistan

**Sub-Theme:** Assessment and evaluation

**Abstract**
In any educational system, no one can deny the importance of assessments. Assessments help in evaluating the knowledge gained by a learner at any specific point as well as in continuous improvement of the curriculum design and the whole learning process. However, with the increase in students' enrollment at University level in either conventional or distance education environment, traditional ways of assessing students' work are becoming insufficient in terms of both time and effort. In distance education environment, such assessments become additionally more challenging in terms of hefty remuneration for hiring large number of tutors. The availability of automated tools to assist the evaluation of students' work and providing students with appropriate and timely feedback can really help in overcoming these problems. We believe that building such tools for assessing students' work for all kinds of courses in not yet possible. However, courses that involve some formal language of expression can be automated, such as, programming courses in Computer Science (CS) discipline.

Learning how to program is the core of CS discipline. It is also becoming an important part even in various academic disciplines. Programming skills can only be learned through practice. Instructors provide various practical exercises to students as assignments to build these skills. Usually, instructors manually grade and provide feedbacks on these assignments. Although in literature, various tools have been reported to automate this process, but most of these tools have been developed by the host institutions themselves for their own use. We at COMSATS Institute of Information Technology, Lahore are conducting a pioneer effort in Pakistan to automate the marking of assignments of introductory programming courses that involve C or C++ languages with the capability of associating appropriate feedbacks for students. In this paper, we basically identify different components that we believe are necessary in building an effective automated assessment system in the context of introductory programming courses that involve C/C++ programming. We further provide a survey of existing state of the art tools and techniques reported in literature for implementing these components. We also discuss those reported concepts and techniques that can help in making such system reusable with the capability of sharing its assessment objects in a secured manner.

**Keywords:** Assessment, programming courses, C/C++

## 1. Introduction

Assessment is an essential part of any educational environment. There are broadly two types of assessments, i.e., formative and summative. Both types of assessments are considered to be necessary for improving curriculum design and the whole learning process. At higher education institutions (especially that follow semester system) that belong to either conventional or distance education environment, teachers performs various tasks to achieve both formative and summative assessments. These tasks typically involve preparing and giving quizzes, assignments, projects, and conduct periodic exams. However, due to increasing number of students enrolling in these institutions, teachers remain under burden during whole semesters. In distance education environment, such assessments become additionally more challenging in terms of hefty remuneration for hiring large number of tutors. Besides assessments teachers also perform other essential tasks such as: preparing course content, conducting lectures, conducting meetings with students to resolve their problems related to course, and exams invigilation. Moreover, they also strive to conduct research for their future prospects and for the common good of the society. Having mentioned the range and nature of tasks performed by a teacher, it is highly probable that a teacher's time and effort might be compromised in performing any or all of these activities.

From the students' perspective, they require quality in lectures and learning materials. They also require fair grading in assessments. They also need comprehensive, immediate, and continuous feedbacks on their work from their respective teachers to improve their learned knowledge and skills. The non-availability of appropriate time and effort from a teacher in performing these tasks can seriously degrade students' learning motivations and quality of learning. In most of the advanced countries, teachers take the help of paid teaching assistants (TAs) to help them in conducting various tasks. However, in developing countries, there is a lack of skilled people and resources to hire such TAs.

We believe that it is time to come up with new ideas to automate various teaching tasks to support effective teaching and students' learning. One task that can be automated is assignments checking and grading. Such automated assessments are usually called as Computer Aided Assessment (CAA). Although, assignments for all types of courses cannot be checked automatically, but courses that have some strict language of expression can be automated, for example, programming courses in the Computer Science (CS) domain.

Programming is an essential part of CS and in even other academic disciplines. Practice is the only key to learn programming for which teachers provide assignments carrying programming problems to be solved. Usually, instructors manually grade and provide feedbacks on these assignments. This is very time consuming and requires a lot of efforts. In distance education environment, peer assessment are also being utilized to deal with this problem but again dealing with conflict of interest situation between students might not be easy every time. In literature, various tools have been reported to automate this process, but most of these tools have been developed by the host institutions themselves for their own use. There are also two open source tools for CAA for programming assignments, namely BOSS[1] and Web-Cat[2]. However, they are monolithic applications and do not support easy integration with other applications such as learning management systems (LMS) (Amelung et al., 2011), authoring tools, and content management systems. As discovered from their respective web sources, they have been originally developed for Java language. In our experience their adaptation for other languages will require a good amount of customization efforts. We at COMSATS Institute of Information Technology, Lahore are conducting a pioneer effort in Pakistan to develop a CAA system for marking of assignments of introductory programming courses that involve C/C++ languages with the capability of associating appropriate feedbacks for students.

In this paper, we basically identify different components that we believe are necessary in building an effective CAA system in the context of introductory programming courses that involve C/C++ languages (see Section 2 for details). The survey is different from conventional surveys where usually developed tools and techniques are only reported. In this paper we also provide a survey of existing state of the art tools and techniques reported in literature for developing/implementing these components. We also report an important software architecture technique that can make assessment system flexible for change and reusable for other external applications (see Section 3). We highlight some standards that have been evolved over the past few years that can help in standardized searching and sharing of assessment objects to support collaboration between teachers and even different systems (Section 4). Moreover, we also highlight reported techniques that can help in securing such sharing of objects (Section 5). It is hoped that this survey will provide a good starting point for those who are interested in building such automated systems.

## 2. Components for Developing an Effective CAA System for C/C++ Programming Assignment

In this section we describe about various important components that are necessary for building an effective CAA system for C/C++ programming assignments provided in introductory programming course. All these components can generate valuable feedbacks for students that can be delivered to students along with teacher's comments.

### A. Program Correctness Evaluation

It is very important for assessment tool to check whether the student's program performs the required functionality (Ala-Mutka, 2005). This is done by comparing the output of the program with the test cases provided by teacher (Ala-Mutka, 2005). The manual generation of test data is very tedious. Although it is not possible to test all inputs for a given program, but there are approaches that try to minimize inputs by using different input variations to identify maximum errors (Myers, 2004). Two well-known approaches for program evaluation are "Equivalence Class Partitioning" (ECP) and "Boundary Value Analysis" (BVA). In ECP, inputs are partitioned into two disjoint classes, i.e., valid and invalid values (Myers, 2004) whereas in BVA, the upper and lower boundaries of input are used as test data (Jorgensen, 2002), (Myers, 2004). From experience, it has been learned that most of errors occur at the boundaries and for values that exist above and below these boundaries (Myers, 2004). The test data generated for both techniques have been described in table 1.

---
[1] http://www.dcs.warwick.ac.uk/boss/about.php
[2] http://wiki.web-cat.org/WCWiki/WhatIsWebCat?action=show&redirect=Web-CAT

Table 1: Examples of Test Data Generated with ECP and BVA Techniques

| For Input | Technique | Possible Test data |
|---|---|---|
| From 1 to 10 | ECP | invalid classes→11, 0, valid class→5 adapted from (Myers, 2004) |
| | BVA | 0, 1, 2, 10, 9, 11 adapted from (Jorgensen, 2002) and (Myers, 2004) |

In practice, both techniques are used together (Jorgensen, 2002). Unfortunately, we could not found any open source tool that can generate such data. For CAA systems one needs to implement these techniques himself. The problem of generating test data becomes complex when more than one input variables are required. The test data for variables can be generated using both BVA and ECP. However, to test a program thoroughly we need all possible combinations of test data values which can be quite large and not feasible for assessment system. However, it has been observed by researchers that most of the errors or bugs in software occur usually due to combination of some variables (up to 6-way combination) (Leung, H., 2014). Therefore, as reported by Leung, test cases can be generated by using the combinations of subsets of all variables. There are various open source tools for addressing test cases combinatorial problem for example, PICT[3], Jenny[4], and AllPairs[5].

**B. Static Analysis of Code**

The static analysis of code helps in identifying those errors that cannot be detected by typical compilers can remain unnoticed even after several executions. However, identification of these errors is necessary as they can create problems at any time in the long run. These errors include, memory leaks, unidentified infinite loops, out of boundary, etc. To enforce good programming practices these problems need to be highlighted for programming students as errors or warnings. Similarly, teachers may use such errors reporting for awarding appropriate marks to students. There are three open source tools that can help in highlighting these errors in the context of C language. These tools are Splint (Secure Programming-Lint)[6], Memwatch[7], and CppCheck[8]. However, it must be noted that CppCheck is the only tool that also supports C++. There are many checks that are supported by these tools. However, it will not be appropriate to report about all of them for students of introductory programming course. The list of checks that might be sufficient is listed in table 2. The table also list tools that support them. One more point that we want to highlight is that some parts of error messages produced by these tools might not be of any interest or understandable for students. Therefore, we can simplify or remove such unnecessary information in these messages.

---

[3] http://msdn.microsoft.com/en-us/library/cc150619.aspx
[4] http://burtleburtle.net/bob/math/jenny.html
[5] http://engineering.meta-comm.com/allpairs.aspx
[6] http://www.splint.org/manual/manual.html
[7] http://www.linkdata.se/sourcecode/memwatch/
[8] http://cppcheck.sourceforge.net/

Table 2: List of Useful Checks and their Support in Static Code Analyzers

| Check | Tool | | |
|---|---|---|---|
| | **Splint** | **Memwatch** | **CppCheck** |
| Memory leak | yes | yes | yes |
| Dangling reference | yes | no | yes |
| Infinite loop | yes | no | no |
| Boundary check | yes | no | yes |
| Unreachable code | yes | no | no |
| Incomplete switch | yes | no | no |
| Return values that never used | yes | no | no |
| Variables not used | yes | no | yes |
| Functions not used | no | no | yes |

**C.  Grading on the Basis of Semantic Similarity with Model Solution(s)**

Similarity checking of students solutions with model solution(s) is a key task in assessment for grading and providing feedback. It can help in grading a program on the basis that whether it meets some design specifications (for example, modularity) (Vujosevic-Janicicet et al., 2012). It can help in evaluating programs that are incomplete or execute infinitely (Tiantian et al., 2007) or unable to compile/execute. Thus this approach can be very valuable in addition to other two modes of assessments, i.e., functionality testing and static analysis of code (Vujosevic-Janicicet et al., 2012) as described in previous sections. In this context, an important work among others has been done in (Vujosevic-Janicicet et al., 2012). The authors used the concept of "Control Flow Graph" (CFG) which represents program's structural flow. According to authors, the graph consists of nodes where each node represents a sequence of code. The authors described that it does not include control and iteration statements. They argued that a CFG treats control flow statements differently from other blocks of code (to determine nodes topology) this is why it is suitable to determine similarities of any two solutions. They used "neighbor matching" algorithm for identification of similarities. They argued that this algorithm possesses such capabilities that are very valuable in comparing student's solution with model solution.

Besides graph based approach, in our opinion, the token matching approach described in section 1.E can also be explored for determining semantic similarity of solutions.

**D.  Programming Style Checking**

The coding style of programmers plays a vital role in software development and its maintenance. Different institutions and projects follow either already available or develop their own homegrown coding standards. Without such standards it becomes hard to understand or change the code by other programmers and even in some cases by the original creators themselves. Style of programming is a skill that must be learned by students right from the beginning to become successful in professional lives. Studies in

(Zaidman, 2004) and (Ala-Mutka, 2004) show that students found the teaching of programming standards very valuable. Although some students did not show interest in such standards, e.g. beginners due to complexity (Ala-Mutka, 2004) but indeed its importance has been acknowledged by teachers and professionals (Zaidman, 2004). According to literature, there is no single standard available for programming styles. One of the primary works related to coding standards was done by Dromey. He provided a framework to associate code level attributes to "high level" software "quality attributes" such as: "reliability", "efficiency", "functionality", "maintainability", "reusability" (Dromey, 1995). Later, Uimonen performed similar kind of analysis and linked source code features with "software quality attributes" (as cited in Ala-Mutka et al., 2004). The authors in (Oman & Cook, 1990) provided "taxonomy" of programming styles after conducting extensive literature survey and reviewing code evaluators. Their style classification include: (i) "Typographic style", (ii) "Control structure style", and (iii) "Information structure style".

Researchers have also put efforts in developing automated analyzers to check programming style for different languages. For example, STYLE and CAP tool for PASCAL, PASS-C for C, and STYLE++ for programs written in C++ (Ala-Mutka, 2004). However, it must be noted that PASS-C is a commercial product. Moreover, we could not found the downloadable binaries or source code of Style++. There is one open source tool for Java called "CheckStyle"[9]. We also found one such tool for C and C++ language called "nsiqcppstyle"[10]. This tool has been implemented in python language and provides very few checks. However, both these tools can help in developing scripts and regular expressions to validate any given style check for C and C++ programming languages. A typical approach of implementing these checks would be to parse the "Abstract Syntax Tree" of any given code. Moreover, the understanding of these tools can also help in developing a framework in which teachers can be facilitated to easily modify any programming style guideline check according to their own needs. For example, CheckStyle as mentioned on its website currently supports flexible architecture in which modifications in any rule is performed by editing its configuration XML file.

E. Source Code Plagiarism Detection

The source code plagiarism detection always remained a focal area in programming assignments. The presence of plagiarism in assignments hinders true learning by students and must be discouraged. According to Parker& Hamblen, a plagiarized source code is "a program which has been produced from another program with a small number of routine transformations" (as cited in Clough, 2000). Several studies reported about the type of changes or transformations that can be adopted by students e.g., see (Liu et al., 2006; Prechelt et al., 2002; Bejarno et al., 2012). In literature, there are basically four broad techniques for automated identification of plagiarism in source code. These approaches include: "metrics based techniques" (Mozgovoy, 2006; Whale, 1990), "token matching" (Cosma & Joy, 2012), "graph based matching" (Liu et al., 2006), and "abstract syntax tree based techniques" (Liu et al., 2006). Currently available tools adopt "token matching" approach due to its simplicity, efficiency, and accuracy.

---

[9] http://checkstyle.sourceforge.net/
[10] https://code.google.com/p/nsiqcppstyle/

According to authors in (Cosma & Joy, 2012), in the token matching approach, the source code is first converted into standard set of tokens. They described that the line of codes which belong to same family of instructions are given the same tokens. For example, control structures if-else, switch statements will be given same token name. The tokens of two sets of source codes are then compared to determine their similarity (Cosma & Joy, 2012) using a similarity function (Prechelt et al., 2002). The Plague, Yap3, Sherlock, and JPlag are the famous tools that adopt this technique (Cosma & Joy, 2012). The authors of YAP3 utilized a novel algorithm for token matching called "Greedy String Tiling Algorithm" (Wise, 1996). The purpose of it is to identify statements reordering and conversion of a function in more than one functions type of cheatings (Cosma & Joy, 2012). The efficiency of the tiling algorithm was further enhanced by utilizing the "Karp-Rabin string matching algorithm" (Wise, 1993). Among the tools that are described earlier, JPlag is accessible as web service. The Sherlock system is claimed to be the part of BOSS assessment system[11] but we could not found its source code in the downloaded source code of BOSS. The source code of Yap3 is available for non-commercial use from its website[12]. However, Yap3 provides tokenizer for only C language. To use Yap3 for C++ language, two open source tokenizers can be explored, i.e., Flex[13] and Quex[14].

To pursue plagiarism cases, good visualization techniques for displaying results are also necessary. For example, in JPlg the results are presented in terms of histograms where each histogram represents similarity between two source codes pairs (Prechelt et al., 2002). According to authors in (Prechelt et al., 2002), to further explore the results for more confirmation of plagiarism, the user can view two source codes in "side-by-side comparison" windows where matching lines are highlighted using similar colors. They further described that by clicking hyperlinks near matching lines on either window, the view in other window automatically jump to its corresponding code. Similarly, the box plot technique reported by authors in (Cosma & Joy, 2012) can also be valuable for proving plagiarism cases. According to authors, the technique helps in identifying copy cases by differentiating them from the source codes of rest of the students.

F. Miscellaneous

There are many other important features that can be evaluated in the context of programming assignments for grading and feedback. They may include for instance, lines of code, "memory usage", efficiency in terms of "CPU time" (Ala-Mutka, 2005). Their implementation does not require much programming efforts.

3. Service-Oriented Architecture (SOA) for Flexibility and Reusability of Components

Until recently, most of the software design architectures were monolithic. Monolithic architectures of increasingly complex software design usually impose strong restrictions, for instance, difficult integration with other software tools, less interoperability and reusability of various software components. Service-oriented architecture (SOA) is a promising architecture to solve these problems, providing loose coupling between modules/services, reusability of modules, and easy adaptations in software in response to changes in business workflows. Moreover, it also facilitate in integrating either whole software or its one or more modules with external applications. The authors (Davies & Davis, 2005) describe two assessment systems that use SOA: E-Learning Framework (ELF) (Wilson et al., 2004) and LeAP (Blinco et el., 2004a). ELF is reported to be an ongoing work that also provides support and means to integrate ELF in the existing LMSs "regardless of the technology they are built on". In contrast, LeAP is reported to be the "first large project of its kind" which adopted SOA. Similarly, JORUM+ "is a repository of educational content" in UK, allowing their reuse by exploiting SOA (JORUM, 2014). In the context of this paper, we find the work presented in (Amelung et al., 2011) to be one of the few works that explains adequately the development of SOA for CAA system for programming assignments. We describe it in some detail in the following paragraph.

---

[11] http://www.dcs.warwick.ac.uk/boss/history.php
[12] http://luggage.bcs.uwa.edu.au/~michaelw/YAP.html
[13] http://flex.sourceforge.net/manual/Cxx.html#Cxx
[14] http://quex.sourceforge.net/index.html

In compliance with SOA, the authors in (Amelung et al., 2011) considered various testing solutions as **Backends**, for instance, JUnit for Java and QuickCheck for Haskell programming language. According to authors, these Backends are basically "self-contained web services" provided securely over "standard internet protocol" using "Python's XML-RPC server API". The authors used the term **Frontend** for GUIs that call Backends. They support two GUIs: ECAutoAssessmentBox and light weight Java frontend: Stand-Alone Thin Client. According to them, Frontend performs functionalities such as: (1) "storage of assignments", their solutions and courses (2) configuration of courses and assignments (3) status and statistics of submitted assignments, etc. To make the architecture flexible and to enable loosely coupled integration of Frontends and Backends, the authors used another component named **Spooler**. According to authors, it acts as an intermediary between various Frontends and Backends; providing "uniform access" to Backends. They further described that this Spooler is in fact similar to the printer spooler which: (1) "manages a submission queue" of service requests by Frontends (2) manages the result queue of outputs to be sent back to Frontend generated by the corresponding Backends, etc.

4. Sharing Assessment Questions in Compliance with IMS GLC

"IMS Global Learning Consortium (IMS GLC)" is "a global, nonprofit, member organization that strives to enable the growth and impact of learning technology in the education and corporate learning sectors worldwide" (IMS-Global, 2014a). As reported on its website, since its inception, the consortium has introduced 20 standards in the field of educational technology. After the efforts of several years, the consortium has introduced IMS Common Cartridge (IMS-CC) to "share and reuse" learning objects (IMS-CC, 2014a) and IMS Digital Repositories Interoperability (IMS-DRI) (IMS-DRI, 2014) standard for querying or searching learning objects repository in a standard way (Queiros & Leal, 2012). The implementations detail of these standards for programming exercises as learning objects for automated evaluation purposes has been demonstrated in (Queiros & Leal, 2012). By following these standards, other applications (other evaluation engines, Learning Management Systems) can reuse repository of programming exercises and other metadata related to them, for example, marking scheme, learning objectives, complexity, source code solutions, test cases, etc. Similarly, they can add their own exercises in shared repository. If required, the assessment system then can respond them by providing results of evaluation. However, it must be noted that IMS does not provide any specification related to the response results and any particular assessment systems can use its own format of reply as reported in (Queiros & Leal, 2012). IMS also provide "questions & tests interoperability" (IMS-QTI)[15] standard for sharing questions and their results between different systems (IMS-Global, 2014b). However, it is not practically capable for dealing with programming exercises evaluation requirements, such as test cases, marking scheme, etc. (Leal & Queiros, 2009).

## 5. Secure Sharing of Assessment Objects

Teachers within or across institutions share their expertise, experience and resources with each other for the betterment of students. In case of CAA, sharing of programming exercises and their related metadata, such as, model solutions, marking scheme, test cases, etc. can be very beneficial in raising the standards of education in computer programming. The IMS standards described in previous section are the means to achieve this goal. By utilizing the collective intelligence of teachers, such sharing can also help in creating a repository of open educational resources (OER) for teachers. The development of open educational resources and their impact on education and society is being researched very passionately. In our view, teachers can be facilitated to share these resources in two modes, i.e., protected and public mode or they can keep them private. In protected mode teachers can share these assets with a group of colleagues, friends, or any group they like. In public mode all resources can be viewed by any teacher. However, such sharing must be made secure to ensure access rights of users. The IMS standard itself provides authorization mechanism for IMS CC specifications (IMS-CC, 2014b). Another work that can help in this context has been done in (Ahmad & Whitworth, 2011). By extending this work, we provide a generalized policy rule for implementing secure sharing of resources as follows:

Let $\delta$ consists of two states named as Public (Pb) and Protected (Pr). $\sigma$ computes the access request decision and has two states Allow (a) and Deny (d). $\sigma$ also has two functions, to map T to Pb, and to map T to Pr so that it can decide on the result of a request. If some teacher's policy states that the Public and Protected members can only access such and such course material, then the access control policy $\Lambda$ may follow the following set of rules:
1) If a teacher outside the domain of the owner T' has a mapping to the teacher T, then T' will become the member (M) of domain, else it remains non-member (Nm).
2) Objects (O) belong to T are assigned a security clearance label (L) to show the sensitivity of the object. These labels are used to share material with other teachers in public and protected role.
3) If T' is a member of domain D(T) and requests some object for which (s)he has the clearance, then the request will be allowed.
4) If T' is a member of domain D(T) and requests some object for which (s)he does not has the clearance then the request will be denied.
5) If T' is not a member of domain D(T) and requests for an object, then in this case the request will be ignored.

---
[15] http://www.imsglobal.org/apip/alliance.html

$$\Lambda = \begin{cases} T' \to M/Nm \ \forall \ T' \in \ D(T) \\ \quad \quad 0 \ \to \ L \\ \quad T' \ \to \ L \ (D(T)) \\ M \to a/d_r \ \forall \ r \ \in \sigma \end{cases}$$

The system implementation is done through centralized security kernel –trusted software component which interrupts every system request and decides its outcome. The access control module issues secure tokens to the requestor, so (s)he can access the object. Digitally signed centralized tokens are used to access the requested objects to ensure the integrity of the system.

**Conclusions**

In this paper, we describe the necessity for developing CAA systems to raise the standards of education in both conventional and distance education environment. We reported different components and techniques that can help in building an effective CAA system for programming assignments that involve C/C++ languages. We also identified some techniques and concepts that can help in making such system reusable with the capability of sharing its assessment objects in a secured manner. It is hoped that this paper will provide a good starting point for those interested in building such CAA systems